\pdfoutput=1

\documentclass[11pt]{article}

\usepackage[final]{coling}

\usepackage{times}
\usepackage{latexsym}
\usepackage{amsmath}
\usepackage[T1]{fontenc}

\usepackage[utf8]{inputenc}

\usepackage{microtype}

\usepackage{inconsolata}

\usepackage{graphicx}
\usepackage{overpic}
\usepackage{multirow}
\usepackage{array}

\title{Enhancing the Reasoning Capabilities of Small Language Models \\ via Solution Guidance Fine-Tuning}

\author{First Author \\
  Affiliation / Address line 1 \\
  Affiliation / Address line 2 \\
  Affiliation / Address line 3 \\
  \texttt{email@domain} \\\And
  Second Author \\
  Affiliation / Address line 1 \\
  Affiliation / Address line 2 \\
  Affiliation / Address line 3 \\
  \texttt{email@domain} \\}

\author{Jing Bi, Yuting Wu\thanks{\;Corresponding author.}, Weiwei Xing$^{*}$, Zhenjie Wei\\
School of Software Engineering\\
Beijing Jiaotong University, Beijing, China\\
\texttt{bijing1206@foxmail.com, \{ytwu1, wwxing\}@bjtu.edu.cn, kogler\_s@outlook.com
}}

\begin{document}
\maketitle
\begin{abstract}
Large language models (LLMs) have demonstrated remarkable performance across a wide range of tasks. Advances in prompt engineering and fine-tuning techniques have further enhanced their ability to address complex reasoning challenges. However, these advanced capabilities are often exclusive to models exceeding 100 billion parameters. Although Chain-of-Thought (CoT) fine-tuning methods have been explored for smaller models (under 10 billion parameters), they typically depend on extensive CoT training data, which can introduce inconsistencies and limit effectiveness in low-data settings. To overcome these limitations, this paper introduce a new reasoning strategy Solution Guidance (SG) and a plug-and-play training paradigm Solution-Guidance Fine-Tuning (SGFT) for enhancing the reasoning capabilities of small language models. SG focuses on problem understanding and decomposition at the semantic and logical levels, rather than specific computations, which can effectively improve the SLMs' generalization and reasoning abilities. With only a small amount of SG training data, SGFT can fine-tune a SLM to produce accurate problem-solving guidances, which can then be flexibly fed to any SLM as prompts, enabling it to generate correct answers directly. Experimental results demonstrate that our method significantly improves the performance of SLMs on various reasoning tasks, enhancing both their practicality and efficiency within resource-constrained environments.\footnote{Code and data available at \url{https://github.com/BiJings/SGFT}.}

\end{abstract}

\section{Introduction}

As large language models (LLMs) continue to expand in scale, they demonstrate remarkable proficiency in tasks such as language generation, translation, question answering, and so on. Moreover, they are increasingly being recognized for their potential in addressing more complex challenges, such as reasoning \citep{Yang2022LanguageMA} and mathematical problem-solving \citep{Mishra2022LILAAU}. The application of LLMs to reasoning tasks has garnered significant academic interest \citep{Qiao2022ReasoningWL},  particularly in the domain of mathematics, where reasoning tasks present substantial challenges \citep{Lu2022ASO}. 
As one of the representative techniques for enhancing the reasoning capabilities of LLMs, Chain-of-Thought (CoT) reasoning \citep{Wei2022ChainOT,Kojima2022LargeLM} enables LLMs to deduce answers step-by-step rather than directly providing an answer, thereby producing more accurate and reliable results. 
However, directly applying CoT reasoning to small language models (SLMs) with fewer than 10 billion parameters has proven to be considerably less effective \citep{Ho2022LargeLM}. \citet{Lanham2023MeasuringFI} propose that CoT prompting only performs effectively under specific scenarios and model scales.

Further, the techniques of CoT fine-tuning have been proposed for SLMs \citep{Ho2022LargeLM}. This kind of methods involves fine-tuning SLMs using CoT data, allowing them to reason through intermediate steps before reaching a conclusion. However, CoT follows a completely independent process for each problem, integrating both logical reasoning and computation. Due to insufficient fitting with a small amount of data during model training, a large amount of CoT training data is required. Acquiring and annotating such data is labor-intensive \citep{LI2022ExplanationsFL}, and the logical and grammatical consistency of the data, which depends on manual annotation, are not always assured.

\begin{figure}[t]
  \centering
  \includegraphics[width=0.9\columnwidth]{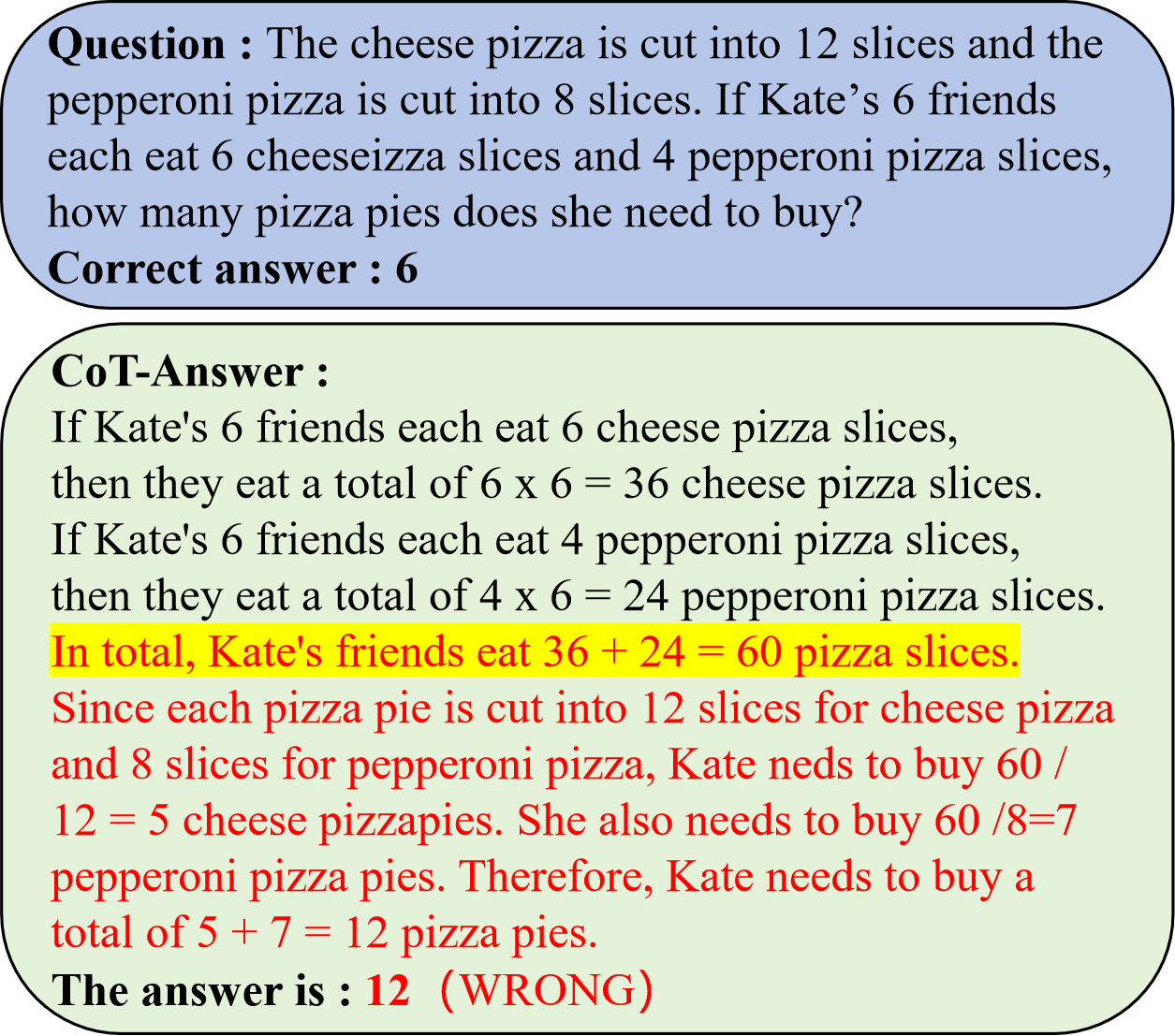}
  \vspace{-1mm}
  \caption{A wrong case of existing CoT reasoning.}
  \label{fig:wrongcase}
\end{figure}

Additionally, existing CoT involves problem-solving steps and a final answer. Each step contains specific explanations and calculations, and the outcome of the current step directly influences the logical generation of the next. This cascading effect causes multiple steps of explanations and calculations to accumulate, leading to error propagation and often generating excessive text, which introduces noise and ultimately affects the accuracy of the final answer. For example, when using CoT reasoning to solve the math problem in Figure~\ref{fig:wrongcase}, an unnecessary step was incorrectly generated during the reasoning process: \emph{In total, Kate's friends eat 36+24=60 pizza slices}, which led to miscalculating the number of cheese pizzas and pepperoni pizzas in the subsequent steps, ultimately resulting in an incorrect final answer. This issue is particularly pronounced in SLMs. Moreover, language models trained with a large amount of CoT data may experience a decline in their equally important general capabilities \citep{Fu2023SpecializingSL}. When faced with common-sense or simple questions, the model may fabricate non-existent reasoning chains, unnecessarily complicating the problem and providing incorrect answers.

To address the above issues of existing CoT techniques, we propose a new training paradigm, \emph{Solution-Guidance Fine-Tuning} (SGFT), to improve the reasoning capabilities of small language models. Different from existing CoT, we present a new reasoning strategy \emph{Solution Guidance} (SG), which only expects the SLMs to generate problem-solving guidance without calculations or extra explanations. SG focuses on problem understanding and decomposition at the semantic and logical levels, rather than specific computations. With only a small amount of SG training data, SGFT can effectively fine-tune a SLM to generate accurate problem-solving guidances. Then, we only need to use the generated solution guidances as input prompts for another SLM, which can produce the correct answers without additional training.

To verify the effectiveness of our method, we conducted experiments on multiple reasoning benchmark datasets. The results demonstrate that, compared to traditional CoT fine-tuning applied directly to SLMs, our method significantly enhances performance in mathematical and common-sense reasoning tasks while preserving the original capabilities of SLMs. Additionally, since our approach requires significantly less training data, it is more practical and efficient for real-world applications.

Our key technical contributions are as follows:

\begin{itemize}
    \item We introduce a new reasoning strategy Solution Guidance (SG) for small language models (SLMs), which focuses on problem understanding rather than specific calculations. SG can effectively reduce data generation costs and significantly improve the reasoning capabilities of SLMs.
    \item We propose a plug-and-play fine-tuning paradigm, named SGFT, for enhancing the reasoning capabilities of SLMs. Using only a small amount of SG training data, SGFT can fine-tune a SLM to generate accurate problem-solving guidances. These guidances can then be flexibly used as input prompts for any SLM, allowing it to directly produce correct answers.
    \item Experiments on multiple reasoning benchmarks datasets demonstrate that our method significantly improve the reasoning capabilities of various SLMs. Our approach can be implemented on a single consumer-grade GPU, and achieve better performance with only 3\% of the training data required by CoT fine-tuning.
  \end{itemize}

\section{Related Work}

\paragraph{\textbf{Chain-of-Thought Reasoning.}}
In prompt engineering, Chain-of-Thought (CoT) \citep{Wei2022ChainOT} has shown excellent performance in reasoning tasks for large models. The method of automatically generating CoT prompts \citep{Zhang2022AutomaticCO} uses similarity-based retrieval methods, improving the performance of large language models in zero-shot reasoning tasks. \citet{Diao2023ActivePW} combine active prompting with CoT, incorporating uncertainty measures and self-consistency methods to enhance the accuracy and consistency of reasoning.  \citet{Turpin2023LanguageMD} believe that the chain of thought explanation is sometimes unreasonable. \citet{Wang2023BoostingLM} introduce the knowledge chain prompting method, which boosts the reasoning abilities of language models, and proposed a post-validation mechanism to ensure the accuracy of the reasoning chain. \citet{Wang2022SelfConsistencyIC} propose to replace greedy decoding with self-consistency, which further improves the effectiveness of CoT. Analyzing incorrect answers in-depth \citep{Zhong2024AchievingO}, by adjusting the CoT for erroneous cases, encourages large language models to better understand the problems and utilize critical information for improved reasoning. \citet{Chae2024LanguageMA} decompose the reasoning process of language modeling into two steps Think-and-Execute improves the reasoning. Thought-of-Tree (ToT) \citep{Yao2023TreeOT} builds on the Chain of Thought approach for more complex planning of tasks.

\paragraph{\textbf{Distillation of Reasoning Ability.}}
Small models have much smaller internal mapping space compared to large models. Despite this, small models still possess certain reasoning abilities \citep{Fu2023SpecializingSL}. Our focus is on leveraging these abilities to solve complex problems. \citet{Ho2022LargeLM} first propose using large-scale language models (such as GPT-3 175B) as teacher models, generating CoT to fine-tune smaller student models. This approach significantly reduces the model size requirements and greatly enhances the performance of small models on complex reasoning tasks. Subsequent work has further refined the CoT fine-tuning method for small models \citep{Magister2022TeachingSL}. \citet{LI2022ExplanationsFL} propose a multi-task learning framework to enable small models to acquire reasoning abilities and generate explanations. \citet{Fu2023SpecializingSL} specialize the training of small models, enhancing multi-step mathematical reasoning tasks. Researchers \citep{Choi2024CanOL} demonstrate that fine-tuned small models can achieve results comparable to large language models in specific domains. This has significant implications for resource optimization in practical applications. Article \citep{Zhu2023ASO} provide a summary of quantization, distillation, and other methods applied to large language models.

\section{Method}

\begin{figure*}[t]
\centering
  \includegraphics[width=0.8\linewidth]{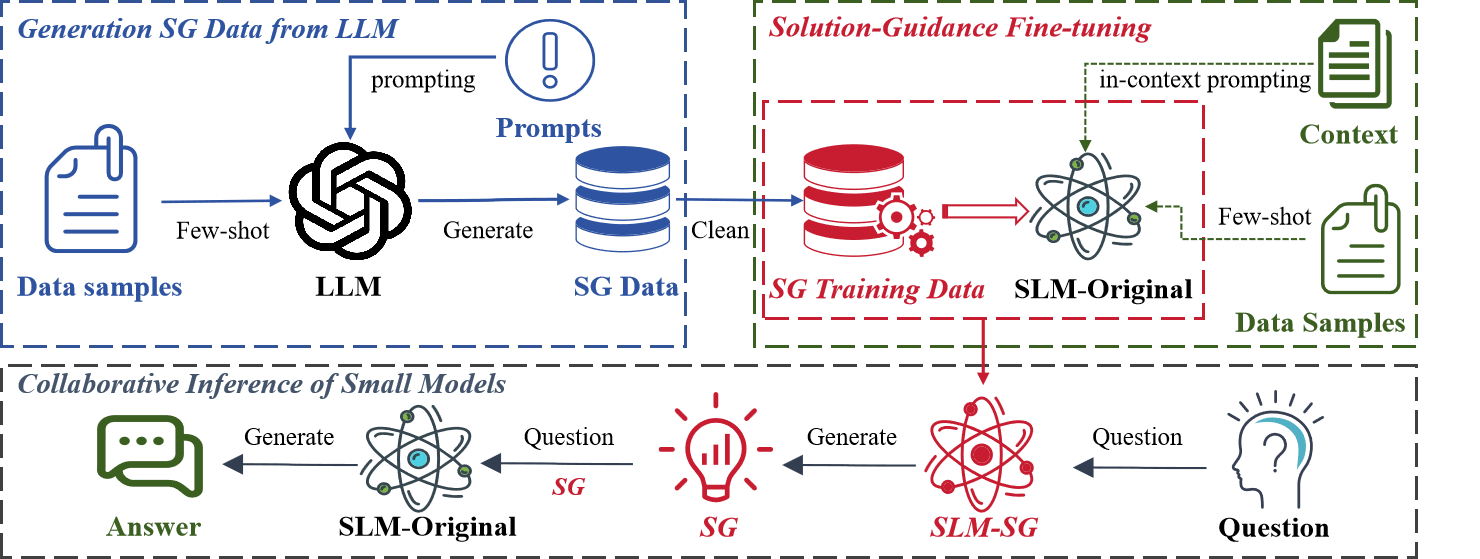} \hfill
    \caption{The figure illustrates the complete architecture of our approach. In this diagram, SLM denotes the Small Language Model, LLM represents the Large Language Model, and SG (Solution Guidance) refers to the data schema introduced in our method.
    }\label{fig:framework}
\end{figure*}

As Figure~\ref{fig:framework} shows, our approach consists of the following steps: solution guidance generation, solution-guidance fine-tuning, and collaborative inference of SLMs based on SG. Specifically, GPT-4o is employed as the teacher model to assist in solution guidance generation. With the generated SG data, we fine-tune a SLM to break down complex problems into a series of manageable solution steps. Then, the generated solution guidances, as well as the original question, are passed to another SLM to generate the final answer without extra training.

\subsection{Solution Guidance Generation}

\begin{table*}
  \centering
  \small
\begin{tabular}{l|l}
\hline
\begin{tabular}[c]{@{}l@{}}Prompts for\\ original SG data\end{tabular} &
  \begin{tabular}[c]{@{}l@{}}\textbf{prompt\_i:} Please generate the {[}Problem Type{]}, {[}Solution Objective{]}, \\ {[}Constraints{]}, and {[}Priorities and Considerations{]} for this question:"Q". \\ \textbf{prompt\_ii:} Please generate a step-by-step solution for this problem. \\ You don't need to solve it; just output the steps in 2 to 6 steps.\end{tabular} \\ \hline
\begin{tabular}[c]{@{}l@{}}Prompts for\\ SG training data\end{tabular} &
  \begin{tabular}[c]{@{}l@{}}\textbf{prompt\_iii:} Please generate a step-by-step solution for this problem. \\ You don't need to solve it, just output the steps in 2 to 6 steps. \\ Examples for the SG data are as follows : {[}Q1+SG1{]}...{[}Q7+SG7{]}\end{tabular} \\ \hline
\begin{tabular}[c]{@{}l@{}}Prompts for \\ finetuning on SLM\end{tabular} &
  \begin{tabular}[c]{@{}l@{}}\textbf{prompt\_iv:} Please generate a step-by-step solution for the following \\ problem with no calculations. You don't need to solve it, just \\ output the steps  in 2 to 6 steps. \\ \textbf{prompt\_v:} Please generate a step-by-step solution for the following \\ problem with no calculations. You don't need to solve it, just output \\ the steps in 2 to 6 steps. Examples for the SG data are as follows : \\ {[}Q1+SG1{]}...{[}Q3+SG3{]} \end{tabular} \\ \hline
\end{tabular}%
  \caption{\label{prompts}
   The table presents the prompts used in different stages. We employed  \textit{prompt\_i} and \textit{prompt\_ii} to generate individual pieces of data from GPT-4o. Subsequently, we used \textit{prompt\_iii} to generate SG data in batches. \textit{Prompt\_iv} and \textit{prompt\_v} were selected as the prompts for adopting the context strategy and the few-shot approach during fine-tuning.
  }
\end{table*}

In fine-tuning large language models, the quality and structure of training data are critical. While traditional data such as CoT has proven effective for many tasks, it still presents certain limitations. This chapter introduces a new type of training data SG, and the method for its generation. Using GPT-4 as the teacher model, we generate data that differs from traditional CoT. We will examine the principles, implementation details, and advantages of this approach in terms of reducing noise and improving data quality.

\paragraph{\textbf{Data Paradigm Setup.}}
First, we randomly selected a subset of 2,000 questions from the GSM8K training set, referred to as \(Q=\left\{q_{i}\right\}^{N}\), and the corresponding Solution Guidance for these questions, denoted as \(d_{i}\), \(D=\left\{d_{i}\right\}^{N}\) forming the set SG. Therefore, \(G=\left\{\left(q_{i}, d_{i}\right)\right\}^{N}\) is a datasets containing N training instances. We conducted a detailed error analysis of CoT method, found that incorrect answers mainly stem from inaccurate solution objectives, calculation errors, and logical flaws. To address these, we experimented and identified that correct step decomposition and sequencing are crucial for high-quality LM responses. Based on this, we defined SG’s structure, which contains step-by-step solution objectives and their sequence. Then, we prompted GPT-4o to generate more SG training data based on this structure.

Our goal is to identify one or more recurring patterns across different questions. Rather than meticulously analyzing and solving each question individually, we extract commonalities in mathematical reasoning problems to derive general solution steps. The fine-tuned model is then used to analyze each question specifically. For example, a typical pattern might be: "Step 1: Identify the primary problem, Step 2: Determine the necessary operations on specific values, Step 3: ...". The fine-tuning task focuses on learning this process, with the procedural patterns also originating from GPT-4.

\paragraph{\textbf{Generation SG Data from LLM.}}
The Solution Guidance (SG) we introduce is a novel concept. For the questions in the training set, we form a set of questions \(Q=\left\{q_{i}\right\}^{M}\), and input each question into GPT-4o, prompting it to output the process of solving the problem. At this point, GPT-4o outputs \textit{M} processes \(e_{i}\) , and answers \(a_{i}\) (in our experiments, we used \textit{M} = 7). After verifying the correctness of the answers, we denote \(E=\left\{\left(q_{i}, e_{i}, a_{i}\right)\right\}^{M}\). Table~\ref{prompts} presents the prompts that we utilized, we employed \textit{prompt\_i} and \textit{prompt\_ii} to obtain the original SG.

We then examine whether these processes \(e_{i}\), exhibit any consistency or regularity and summarize the patterns identified. Based on the responses, we outline the problem-solving pattern as follows: "1. Identify the total requirements and known information; 2. Solve step-by-step, setting intermediate goals and calculating their values; 3. Summarize all intermediate results and perform unit conversions; 4. Arrive at the final answer and verify it." Subsequently, we used \textit{prompt\_iii} in the Table~\ref{prompts} to generate the SG for question \(q_{i}\) based on this pattern as \(d_{i}\). Examples of the data are shown in the Table~\ref{examples}, resulting in \(G=\left\{\left(q_{i}, d_{i}\right)\right\}^{N}\) as the training data.

It is evident that our SG differs from the original CoT approach. While CoT emphasizes a step-by-step solution, including specific calculations for each problem, SG offers higher-level guidance. SG focuses on providing a framework for problem-solving rather than detailing the exact computational process.

\paragraph{\textbf{Data Cleaning}.}In the generated datasets, we manually replaced some instances involving specific numerical calculations to ensure that the datasets focuses on problem understanding and textual logical reasoning. We also removed data related to pure mathematical calculations, as numerical computation is not the capability we aim to generalize.

\begin{table*}
  \centering
  \small
\begin{tabular}{ll}
\hline
Question& Solution Guidance\\ \hline
\begin{tabular}[c]{@{}l@{}}Bailey starts with a certain amount of money.\\  Then she receives a weekly allowance of \$5 \\ for 8,00 weeks.At the end of the 8,00 weeks,\\  if she has a total of \$100,how much money \\ did Bailey start with?\end{tabular}                                                                                                                      & \begin{tabular}[c]{@{}l@{}}Step 1: Calculate the total amount of \\ allowance Bailey receives in 8 weeks.\\ Step 2: Calculate the amount of money\\  Bailey started with.\end{tabular}                                                                                                                                                                  \\ \hline
\begin{tabular}[c]{@{}l@{}}A classroom has a whiteboard which is shared \\ between the 4 teachers who take turns using\\  the classroom. Each teacher has 2 lessons per \\ day and uses the whiteboard in each lesson.\\  If the whiteboard is cleaned 3 times per lesson,\\  how many times is the whiteboard cleaned in a day?\end{tabular}                                     & \begin{tabular}[c]{@{}l@{}}Step 1: Calculate the total number of\\  lessons in a day. \\ Step 2: Calculate the total number of times\\  the whiteboard is cleaned in a day.\end{tabular}                                                                                                                                                                \\ \hline
\begin{tabular}[c]{@{}l@{}}An interior design firm offers installation for \\ \$129.00. It includes hanging 4 mirrors,\\  2 shelves, 1 chandelier, and 10 pictures.\\ They will install additional items for an extra\\  \$15.00 per item. Angela has 6 mirrors and\\  2 chandeliers and 20 pictures that she needs \\ installed/hung. How much will this cost her?\end{tabular} & \begin{tabular}[c]{@{}l@{}}Step 1: Determine the base installation cost.\\ Step 2: Determine the number of items included\\ in the base installation service.\\ Step 3: Calculate the number of additional items \\ Angela needs installed. \\ Step 4: Calculate the cost for the additional items.\\ Step 5: Calculate Angela's total cost.\end{tabular} \\ \hline
\end{tabular}
  \caption{\label{examples}
    Examples of training data paradigms in our approach.
  }
\end{table*}

\subsection{Solution-Guidance Fine-Tuning (SGFT)}

After acquiring the SG data, we focus on transferring the data generation capability to smaller models. We explore fine-tuning strategies and analyze the effects of incorporating contextual information, as well as zero-shot and few-shot examples, on model performance. Notably, our fine-tuning approaches and reasoning tasks can be executed on consumer-grade GPUs.

\paragraph{\textbf{Fine-Tuning Method Based on LISA.}}

Recent work by \citet{Pan2024LISALI} revealed that LoRA's \citep{Hu2021LoRALA} fine-tuning primarily targets the lowest and highest layers of large language models (LLMs), specifically the embedding and linear head layers. Building on this insight, the LISA method was proposed, which significantly reduces GPU memory consumption by selectively freezing certain layers using layer-wise importance sampling during training. For a 7B model, LISA demonstrates a training speed approximately 1.5 times faster than LoRA.

LISA employs a regularization loss function based on the AdamW optimizer, ensuring training stability while effectively leveraging the layer-wise importance sampling strategy. The objective function optimized by LISA includes a regularization term and is formulated as follows:

\begin{equation}
  \label{eq:loss function}
  f_{\text {reg }}(w)=f(w)+\frac{1}{2} w^{T} S w
\end{equation}

In this context, \(S\) is a finite positive semi-definite diagonal matrix, and \(w\) represents the model parameters. This regularization term helps control the model's complexity and prevents overfitting. 
LISA has the following convergence guarantee during optimization, where \(f_{\text {reg }}^{*}\) represents the optimal value of \(f_{\text {reg }}\):

\begin{equation}
  \label{eq:convergence guarantee}
  \frac{1}{T} \sum_{t=1}^{T} f_{\text {reg }}\left(w_{t}\right)-f_{\text {reg }}^{*} \leq O\left(\frac{1}{\sqrt{T}}\right)
\end{equation}

To optimize fine-tuning efficiency and balance computational resources, we adopted this advanced fine-tuning method. By adjusting only a small number of parameters, this approach enables effective fine-tuning even with limited computational resources.

\begin{table*}
  \centering
  \small
\begin{tabular}{lccccc}
\hline
\textbf{Model}             & \textbf{GSM8K} & \textbf{SVAMP} & \textbf{MultiArith} & \textbf{StrategyQA} & \textbf{CommonsenseQA} \\ \hline
ChatGLM3-6B& 27.4           & 40.8           & 53.8                & 62.3                & 58.1                   \\
ChatGLM3-6B\_CoT            & 34.4           & 47.5           & 62.1                & 69.5                & 64.9                   \\
ChatGLM3-6B\_SG+ChatGLM3-6B & \textbf{43.7}           & \textbf{51.2}           & \textbf{67.4}                & \textbf{72.7}                & \textbf{69.8}                   \\ \hline
Qwen2-7B                   & 23.6           & 32.4           & 50.2                & 64.1                & 59.2                   \\
Qwen2-7B\_CoT               & 31.2           & 37.2           & 57.2                & 70.3                & 64.4                   \\
Qwen2-7B\_SG+Qwen2-7B       & \textbf{39.4}           & \textbf{45.7}           & \textbf{61.4}                & \textbf{72.3}                & \textbf{68.7}                   \\ \hline
Llama2-7B                  & 13.3           & 38.0           & 48.3                & 56.4                & 52.9                   \\
Llama2-7B\_CoT              & 21.8           & 43.4           & 52.6                & 60.6                & 57.3                   \\
Llama2-7B\_SG+Llama2-7B     & \textbf{28.6}           & \textbf{47.8}           & \textbf{57.4}                & \textbf{64.8}                & \textbf{62.1}                   \\ \hline
Llama2-7B\_SG+ChatGLM3-6B   & 38.1           & 46.6           & 61.9                & 69.4                & 64.3                   \\
Llama2-7B\_SG+Qwen2-7B      & 35.4           & 43.8           & 58.2                & 66.2                & 62.7                   \\
ChatGLM3-6B\_SG+Llama2-7B   & 39.6           & 47.9           & 64.7                & 71.7                & 68.2                   \\
ChatGLM3-6B\_SG+Qwen2-7B    & 40.2           & 50.3           & 65.4                & 70.2                & 67.4                   \\
Qwen2-7B\_SG+Llama2-7B      & 42.7           & 49.1           & 67.8                & 73.6                & 69.3                   \\
Qwen2-7B\_SG+ChatGLM3-6B    & \textbf{48.3}  & \textbf{57.8}  & \textbf{72.9}       & \textbf{79.8}       & \textbf{75.7}          \\ \hline
  \end{tabular}
  \caption{\label{main-results}
    We conducted accuracy comparisons across five datasets related to reasoning, using the original SLM version as a baseline. We compared the performance of the original model, the Fine-tune-CoT method (fine-tuned with 30,000 CoT samples), and our method (fine-tuned with 3,000 SG samples). Additionally, we evaluated the models across different smaller-scale versions. All experiments in the comparison table utilized zero-shot prompting.
  }
\end{table*}

\paragraph{\textbf{Improvements in Fine-Tuning Based on Prompt Engineering.}}
Our experiments demonstrated that incorporating context during fine-tuning significantly influenced the model's behavior. For example, when fine-tuning without contextual prompts using only the data \(G=\left\{\left(q_{i}, d_{i}\right)\right\}^{N}\), the model unexpectedly generated specific numerical calculations and final answers, despite this not being the intended output at that stage. By introducing contextual \textit{prompt\_iv} or \textit{prompt\_v} in the Table~\ref{prompts}, the training data was modified to \(G^{\prime}=\left\{\left(\text { prompt }+q_{i}, d_{i}\right)\right\}^{N}\), leading the model to generate problem-solving steps rather than final answers. These findings were validated through our experimental results.

We conducted a series of experiments to assess the impact of different contexts on model performance. Specifically, we tested various contextual configurations \citep{Dong2023ASF}, ranging from short to long contexts, and observed that appropriate context settings significantly improved task completion. In addition, we examined the effects of zero-shot and few-shot examples. Zero-shot tasks were performed without any examples, whereas few-shot tasks included a limited number of examples. Our training data combined task context with sample examples, and the results showed that few-shot examples consistently led to better model performance than zero-shot examples. Due to hardware constraints, the maximum number of examples was limited to three.

\subsection{Collaborative Inference of Small Models}

After fine-tuning the smaller model, our approach enables the model to generate solution guidance rather than directly solving problems. Its role is to parse each input problem and generate the corresponding SG, which is then inputted, along with the original problem, into a response model. This response model can be any generalized language model; In our case, we chose a model with fewer than 10B parameters that had not been fine-tuned for specific tasks, such as a base or chat small model. By leveraging the synergy between our Model\_SG and the language model, we significantly improved accuracy on mathematical and common-sense reasoning datasets compared to the original model.

This division of labor allows the models to retain their general capabilities while efficiently addressing complex reasoning tasks. In our experiments, we compared the performance of an untuned small model serving as the guidance model with a single fine-tuned model responsible for both guidance and answer generation. The results confirmed that our approach, which separates these tasks, is more effective overall.

Our model can be fine-tuned and deployed using a single consumer-grade GPU, yielding superior results compared to traditional Chain-of-Thought (CoT) fine-tuning. This opens up new possibilities for researchers to explore the potential of small language models, even when working with constrained computational resources.

\section{Experiments}

\subsection{Experimental Setup}
\paragraph{\textbf{Training Data:}}
Following the method outlined by \citet{Brown2020LanguageMA}, we randomly selected 7 questions from the GSM8K training set. Using the prompting procedure detailed in the previous, we tasked GPT-4o with generating problem-solving guidance (SG). We then input these 7 questions into GPT-4o to obtain their corresponding SG. Afterward, both the original questions and their SG were fed back into GPT-4o to generate the final answers. Once the answers were verified for accuracy, these 7 questions and their SG were used as few-shot examples for the large model. Then we randomly selected 1000, 2000, and 3000 questions from the GSM8K training set, instructing the large model to generate SG for each. This process provided the necessary training data. We then used the format of original questions plus SG to generate the final answers. Any data with incorrect answers was removed to ensure the training datasets maintained a high level of accuracy.

\begin{figure}[t!]
  \includegraphics[width=0.9\columnwidth]{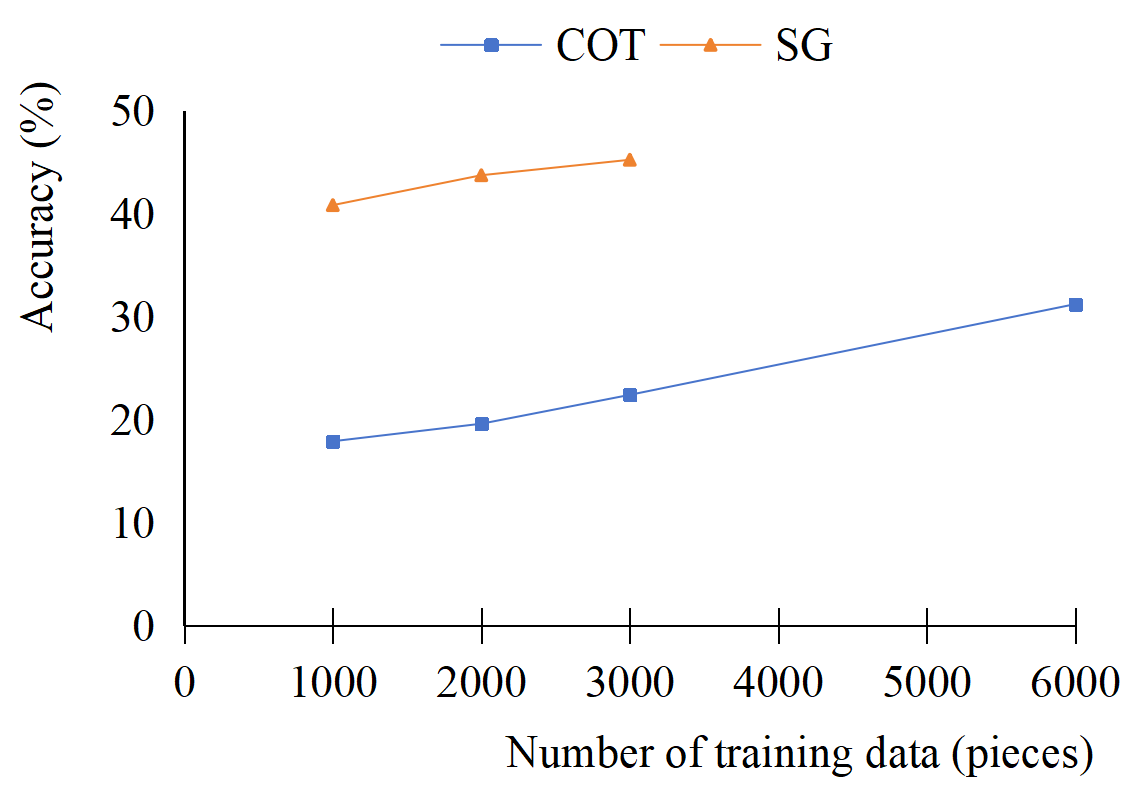}
  \vspace{-3mm}
  \caption{Relationship between the amount of training data required by the CoT method and our method and the accuracy on the GSM8K data set, with the x-axis representing the amount of data and the y-axis representing the accuracy.}
  \label{fig:data}
\end{figure}

\begin{figure}[t!]
\centering
  \includegraphics[width=0.9\columnwidth]{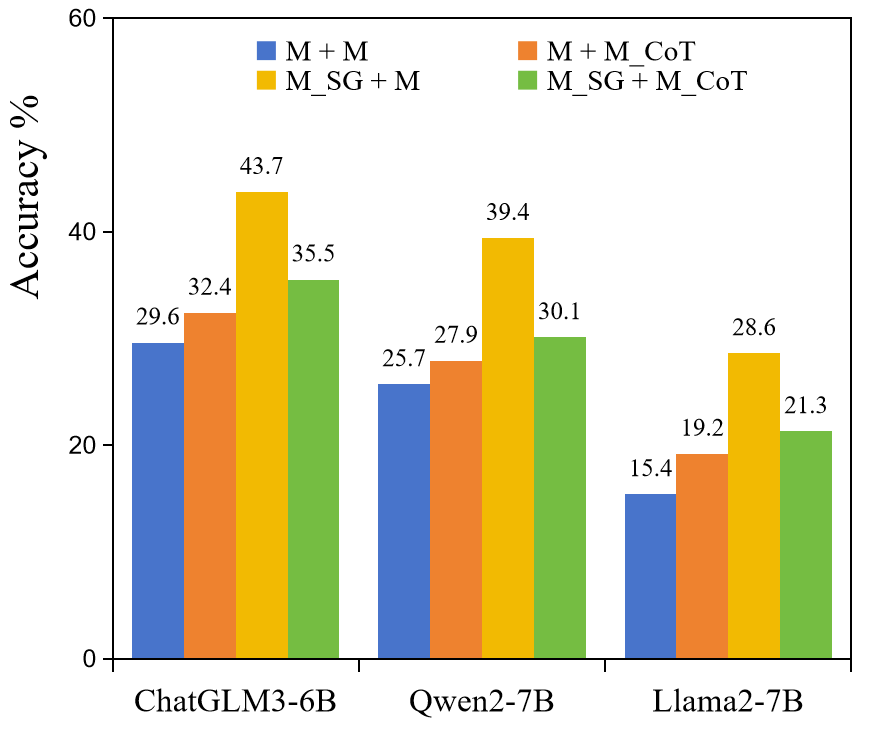}
  \vspace{-3mm}
  \caption{Comparison of the accuracy of four strategies for collaborative reasoning with small models, M represents an original model, M\_CoT represents a model fine-tuned with CoT, and M\_SG represents a model fine-tuned with SG.Test results on the GSM8K dataset.}
  \label{fig:ablation}
\end{figure}

\paragraph{\textbf{Inference Tasks:}}
For mathematical application problems, we used GSM8K \citep{Cobbe2021TrainingVT}, SVAMP \citep{Patel2021AreNM}, and MultiArith \citep{Roy2016SolvingGA}. For commonsense reasoning questions, we used CommonsenseQA \citep{Talmor2019CommonsenseQAAQ} and StrategyQA \citep{Geva2021DidAU}. These datasets are commonly employed to assess the reasoning capabilities of LLMs. Our evaluation metric is the accuracy on each dataset. Notably, GSM8K represents an in-distribution dataset, while the remaining datasets are out-of-distribution, providing a more diverse test of the model's generalization abilities.

\paragraph{\textbf{Language Models:}}
From the perspective of model distillation, we used GPT-4o as the teacher model. The student models included qwen2-7B-Instruct \citep{Chu2024Qwen2AudioTR}, chatglm3-6B-Instruct \citep{Zeng2024ChatGLMAF}, and llama2-7B-chat \citep{Touvron2023Llama2O}. To maintain consistent outputs, the temperature was set to 0 for all models. We performed cross-model inference, using different models to generate SG and final answers. For baselines, we compared with the zero-shot outputs of each instruction-tuned small model.

\subsection{Main Results}


As shown in Table~\ref{main-results}, the collaborative inference method for small models improves their ability to solve complex mathematical reasoning tasks. Across various mathematical and textual reasoning datasets, our approach significantly boosts accuracy. Traditional CoT fine-tuning requires large datasets and tends to focus too narrowly on individual problems, limiting its ability to generalize problem-solving patterns. In contrast, our method uses one model for high-level guidance—understanding the problem and providing strategies—while the other model generates answers based on this guidance. This division of tasks creates an efficient problem-solving process while preserving the general capabilities often lost in fine-tuning.

\begin{table}
  \centering
  \scriptsize
\begin{tabular}{l|ccc}
\hline
\textbf{fine-tuning strategy}                                                        & \multicolumn{1}{l}{\textbf{GSM8K}} & \multicolumn{1}{l}{\textbf{SVAMP}} & \multicolumn{1}{l}{\textbf{MultiArith}} \\ \hline
\begin{tabular}[c]{@{}l@{}}ChatGLM3-6B\_SG*\\ -Zero-shot without context\end{tabular} & 36.4                               & 45.2                               & 55.3                                    \\ \hline
\begin{tabular}[c]{@{}l@{}}ChatGLM3-6B\_SG*\\ -Zero-shot with context\end{tabular}    & 43.7                               & 51.2                               & 67.4                                    \\ \hline
\begin{tabular}[c]{@{}l@{}}ChatGLM3-6B\_SG*\\ -3 shot with context\end{tabular}       & \textbf{48.9}                      & \textbf{57.1}                      & \textbf{70.8}                           \\ \hline
\begin{tabular}[c]{@{}l@{}}Qwen2-7B\_SG*\\ -Zero-shot without context\end{tabular}    & 31.2                               & 38.3                               & 53.7                                    \\ \hline
\begin{tabular}[c]{@{}l@{}}Qwen2-7B\_SG*\\ -Zero-shot with context\end{tabular}       & 39.4                               & 45.7                               & 61.4                                    \\ \hline
\begin{tabular}[c]{@{}l@{}}Qwen2-7B\_SG*\\ -3 shot with context\end{tabular}          & \textbf{46.1}                      & \textbf{50.7}                      & \textbf{68.2}                           \\ \hline
\begin{tabular}[c]{@{}l@{}}Llama2-7B\_SG*\\ -Zero-shot without context\end{tabular}   & 18.8                               & 41.7                               & 51.2                                    \\ \hline
\begin{tabular}[c]{@{}l@{}}Llama2-7B\_SG*\\ -Zero-shot with context\end{tabular}      & 28.6                               & 47.8                               & 57.4                                    \\ \hline
\begin{tabular}[c]{@{}l@{}}Llama2-7B\_SG*\\ -3 shot with context\end{tabular}         & \textbf{37.2}                      & \textbf{52.3}                      & \textbf{63.9}                           \\ \hline
\end{tabular}
  \caption{\label{fine-tuning-strategies}
    We compared the effects of different fine-tuning strategies across three datasets, using the following variables: no context, context with zero samples, and context with 3 samples. SG* refers to the model fine-tuned using SG as the guide model, with the original version of the same model serving as the response model.
  }
\end{table}

\begin{table*}[]
\centering
\scriptsize
\begin{tabular}{ll}
\hline
\multicolumn{2}{l}{\begin{tabular}[c]{@{}>{\raggedright\arraybackslash}p{0.9\linewidth}@{}}Question: The cheese pizza is cut into 12 slices and the pepperoni pizza is cut into 8 slices. If Kate's 6 friends each eat 6 cheese pizza slices and 4 pepperoni pizza slices, how many pizza pies does she need to buy?\\ \textbf{Correct answer: 6}\end{tabular}} \\ \hline
\multicolumn{1}{l|}{CoT} &
  \begin{tabular}[c]{@{}>{\raggedright\arraybackslash}p{0.85\linewidth}@{}}Answer:\\ If Kate's 6 friends each eat 6 cheese pizza slices, then they eat a total of 6 x 6 = 36 cheese pizza slices. If Kate's 6 friends each eat 4 pepperoni pizza slices, then they eat a total of 4 x 6 = 24 pepperoni pizza slices. In total, Kate's friends eat 36 + 24 = 60 pizza slices. Since each pizza pie is cut into 12 slices for cheese pizza and 8 slices for pepperoni pizza, Kate needs to buy 60 / 12 = 5 cheese pizza pies. She also needs to buy 60 /8=7 pepperoni pizza pies. Therefore, Kate needs to buy a total of 5 + 7 = 12 pizza pies.\\ \textbf{The answer is: 12}\end{tabular} \\ \hline
\multicolumn{1}{l|}{\multirow{2}{*}{Ours}} &
  \begin{tabular}[c]{@{}>{\raggedright\arraybackslash}p{0.85\linewidth}@{}}Solution Guidance(SG):\\ Step 1: Calculate the total number of cheese pizza slices needed per person.\\ Step 2: Calculate the total number of pepperoni pizza slices needed per person.\\ Step 3: Calculate the total number of cheese pizzas needed.\\ Step 4: Calculate the total number of pepperoni pizzas needed.\\ Step 5: Sum up the total number of pizzas needed.\end{tabular} \\ \cline{2-2} 
\multicolumn{1}{l|}{} &
  \begin{tabular}[c]{@{}>{\raggedright\arraybackslash}p{0.85\linewidth}@{}}Answer:\\ Step 1: Calculate the total number of cheese pizza slices needed per person.\\ Kate's 6 friends each eat 6 cheese pizza slices, so the total number of cheese pizza slices needed is 6 friends * 6 slices/friend = 36 slices.\\ Step 2: Calculate the total number of pepperoni pizza slices needed per person.\\ Kate's 6 friends each eat 4 pepperoni pizza slices, so the total number of pepperoni pizza slices needed is 6 friends * 4 slices/friend= 24 slices.\\ Step 3: Calculate the total number of cheese pizzas needed.\\ Since each cheese pizza has 12 slices, the total number of cheese pizzas needed is 36 slices / 12 slices/pizza = 3 pizzas.\\ Step 4: Calculate the total number of pepperoni pizzas needed.\\ Since each pepperoni pizza has 8 slices, the total number of pepperoni pizzas needed is 24 slices / 8 slices/pizza = 3 pizzas.\\ Step 5: Sum up the total number of pizzas needed.\\ The total number of pizzas is 3 cheese pizzas + 3 pepperoni pizzas = 6 pizzas.\\ \textbf{The answer is: 6}\end{tabular} \\ \hline
\end{tabular}%
  \caption{\label{case study}
   A test example where CoT fine-tuning method provides an incorrect answer. Our models generated SG and replied with the correct answer based on the SG.
  }
\end{table*}

Traditional CoT methods and our approach differ in the scale of training data. As shown in Figure~\ref{fig:data}, under the same fine-tuning strategy, our method requires only 3.3\% of the data used by CoT to achieve superior performance. This greatly reduces the computational cost for fine-tuning and lowers the barrier for practical application. We compared performance across datasets with 1,000 to 3,000 training examples, noting that accuracy improved with more data. This suggests that our approach has greater potential than traditional CoT methods.

\subsection{Ablation Study}

In this section, we conducted a detailed ablation study to evaluate the effectiveness of fine-tuning models using SG. Consistently following our method, the first model generates SG, while the second model provides the answers. We tested four strategies:
(1) Two untrained models directly for collaborative inference.
(2) One untrained model for guidance and a CoT-fine-tuned model for answering.
(3) One SG-fine-tuned model for guidance and a untrained model for answering.
(4) One SG-fine-tuned model for guidance and a CoT-fine-tuned model for answering.

As shown in Figure~\ref{fig:ablation}, the optimal setup was using an SG-fine-tuned model for guidance and an untrained model for answering. The results suggest that SG-trained models offer better Solution Guidance, leading to more accurate answers from the response model. However, when the CoT model served as the response model, its performance was worse compared to the original model.

\subsection{Comparison among In-context Learning and Few-shot / Zero-shot Learning}

We further explore the effects of in-context (with \textit{prompt\_iv} in the Table~\ref{prompts}), zero-shot, and few-shot examples (3-shot, with  \textit{prompt\_v} in the Table~\ref{prompts}) during fine-tuning on the performance of SG small model collaborative inference.

In Table~\ref{fine-tuning-strategies}, we found that when using our method for fine-tuning, the model's performance was consistently worse with the zero-shot no-context strategy compared to training with context. Training with context yielded better results, and increasing the number of examples further improved performance. However, due to equipment and time constraints, we limited our experiments to a maximum of 3-shot examples. The experimental results indicate that the context plus few-shot strategy is the most effective.

\begin{table*}[]
\centering
\scriptsize
\begin{tabular}{ll}
\hline
\multicolumn{2}{l}{\begin{tabular}[c]{@{}>{\raggedright\arraybackslash}p{0.9\linewidth}@{}}Question: Where is there a telephone book in almost every room ? \\ A). at hotel; B). house; C). library; D). bedsit; (E). closet;\\ \textbf{Correct answer: A}\end{tabular}} \\ \hline
\multicolumn{1}{l|}{\multirow{2}{*}{Ours}} &
  \begin{tabular}[c]{@{}>{\raggedright\arraybackslash}p{0.85\linewidth}@{}}Solution Guidance(SG):\\ Step 1: Analyze the question to understand the context and identify the key concept, which is "a place where a telephone book is commonly found in almost every room."\\ Step 2: Examine the provided choices (hotel, house, library, bedsit, closet) and consider the relevance of each option to the key concept.\\ Step 3: Evaluate each option based on common knowledge about the presence of telephone books in different types of rooms.\\ Step 4: Determine the best answer based on the evaluation and reasoning.\end{tabular} \\ \cline{2-2} 
\multicolumn{1}{l|}{} &
  \begin{tabular}[c]{@{}>{\raggedright\arraybackslash}p{0.85\linewidth}@{}}Answer:\\ Step 1: The question is asking where one would commonly find a telephone book in almost every room. The key concept here is the widespread availability of a telephone book within a certain type of space.\\ Step 2: Let’s examine the choices:\\ A) At hotel: Hotels often provide guests with a telephone book in each room for convenience.\\ B) House: Telephone books are not typically found in every room of a house, but rather in a common area such as a living room or kitchen.\\ C) Library: Libraries do not have telephone books in every room, as they are typically found in reference sections or at the information desk.\\ D) Bedsit: A bedsit, which is a single room that serves as both a bedroom and a sitting room, might have a telephone book, but this is not a guarantee in every bedsit.\\ E) Closet: A closet is not a place where one would expect to find a telephone book.\\Step 3: Based on common knowledge, hotels are known to provide a telephone book in each room as part of their hospitality services. Houses and libraries do not have telephone books in every room, and closets are not places where telephone books are stored. Bedsits, while they could have a telephone book, do not guarantee one in every room.\\ Step 4: The best answer, considering the widespread availability of telephone books in hotels, is A) At hotel.\\ \textbf{The answer is: A}\end{tabular} \\ \hline
\end{tabular}%
  \caption{\label{case study_c}
   A test example from CommonsenseQA.
  }
\end{table*}

\subsection{Case Study}
Table~\ref{case study} presents a test case from the GSM8K dataset where the traditional CoT approach causes the model to generate language logic and perform calculations simultaneously. This results in a misunderstanding of the problem, leading to computational errors and an incorrect answer. In contrast, our method first generates a solution plan (SG) for the problem and then performs calculations based on this plan. This structured approach allows the model to focus on distinct tasks, improving problem logic and ultimately producing a more accurate answer.

Table~\ref{case study_c} shows Our method is also effective in common-sense reasoning problems. SG forms a complete problem-solving path, which includes analyzing the purpose of the problem, the differences among various options, and evaluating each option. Based on the guidance of the SG, the model can obtain the final answer.

\section{Conclusion}
This paper proposes a new plug-and-play training paradigm Solution-Guidance Fine-Tuning (SGFT) for improving the reasoning capabilities of small language models, without relying on large amounts of Chain-of-Thought (CoT) training data. Experimental results demonstrate that the proposed approach significantly enhances SLMs' performance in mathematical and common-sense reasoning tasks while maintaining their general capabilities. A new reasoning strategy Solution Guidance (SG) is introduced, which emphasizes problem understanding and decomposition, reducing noise from calculation steps in traditional CoT data. Experiments show that using 1,000 pieces of SG data outperform 30,000 pieces of CoT data.

\section{Limitations}
In our work, we aimed to enhance the model's reasoning capabilities while preserving its original abilities, so we chose not to focus on smaller-scale models. Some researchers have found that even smaller models, when fine-tuned, can develop certain mathematical reasoning abilities. Future work could explore using smaller models (<3B) as guidance models.
We also observed that many studies employ diverse path reasoning, which can effectively improve answer accuracy. Applying this to our method would involve generating multiple SG for each question, then selecting the final answer based on the most consistent output across all SG. However, given our current computational limitations, we plan to explore these methods further in future work.

\section*{Acknowledgements}
This work is supported by the National Natural Science Foundation of China under Grant 62406022. 

\bibliography{custom}

\appendix

\end{document}